# Towards an efficient inverse static model of a Festo actuator made of two antagonist muscles for hybrid control of its position and stiffness

Bertrand Tondu

Département de Génie Electrique de l'INSA de Toulouse and LAAS/CNRS in Toulouse
bertrand.tondu@insa-toulouse.fr

*Abstract.*

The Festo air muscle is today one of the most known commercial version of the so-called McKibben pneumatic artificial muscle. A major advantage of hand-made McKibben muscles, as well as its commercial versions, lies in the possibility it offers of realizing antagonist muscle actuator on the model of the biceps-triceps system. If pressures are independently controlled in each artificial muscle, it is then possible to define a position-stiffness control of the antagonist actuator by analogy with natural neural control of antagonist skeletal muscles. Such a control however requires a knowledge model of the actuator making possible a stiffness estimation provided by control pressures, while position closed-loop control is facilitated by a feedforward model of this highly nonlinear actuation device. We discuss this issue in the particular case of the antagonist Festo air muscle actuator, and we propose a simplified static actuator model derived from the classic static theoretical model of the McKibben artificial muscle, simple enough for a future MIMO position-stiffness controller be able to integrate it. If the proposed model highlights the ability of the antagonist Festo muscle actuator to mimic the stiffness neural control by the mean of the sum of pressures inside artificial muscles, it also highlights the difficulty to derive a simple but accurate model of the static force produced by Festo muscle in the full control range of pressure. The reported results suggest that it would be particularly interesting to derive advantage from numerous studies about Festo muscle modelling to go further in order to find such an inverse actuator model including an accurate estimation of the actuator stiffness.

## Introduction.

The Festo air muscle can be considered as the most advanced industrial version of the so-called McKibben muscle. Because McKibben artificial muscle is known to have a static and dynamic behavior is analogy with skeletal muscle (Caldwell, Medrano-Cerda & Goodwin, 1995), Chou & Hannaford, 1996), (Tondu & Lopez, 2000), the actuator made of two antagonist artificial muscles is a very interesting candidate for realizing an actuation device with both position and stiffness control. However, if joint position can be measured by means of an encoder, stiffness cannot be directly deduced from a sensor measurement; it must be accurately estimated for an efficient use of such new actuation device. In the framework of this report, we would like to analyze the possibility of determining a formal direct and inverse model of the actuator made of two identical Festo air muscle with the aim of a future position/stiffness MIMO control based on a knowledge model of the actuator static behavior. The report is organized as follows: in a first section, the theoretical static model of the McKibben artificial muscle is recalled in relation with the fundamental phenomenological static model proposed by Neville Hogan for highlighting the stiffness variation of the skeletal muscle. In a second section, we will try to analyze the specificity of the Festo air muscle considered with respect to the McKibben artificial muscle and we will discuss some recent static model of its contraction force before proposing a possible model. In a third section we will apply our proposed Festo muscle static model to the determination of an inverse model of the antagonist Festo muscle actuator.

## 1. The theoretical McKibben artificial muscle in relation with Hogan's phenomenological model of the skeletal muscle.

In his seminal article, Neville Hogan (1984) proposed a simplified static force model of the active force of the skeletal muscle as shown in Fig. 1.a. Such very simplified model has the great advantage to highlight the ability of the skeletal muscle to both generate a force and a stiffness when the neural activity – normalized by the variable $u$ in Hogan's model – varies. This simplified model expressing the static contraction force $F(\varepsilon, u)$ positively counted when the current muscle length $l$ decreases, can be written as follows:

$$F(\varepsilon, u) = uF_{max}\left(1 - \frac{\varepsilon}{\varepsilon_{max}}\right), \quad 0 \leq \varepsilon \leq \varepsilon_{max}, \quad 0 \leq u \leq 1 \tag{1}$$

where $F_{max}$ is the maximum isometric force produced by the muscle when $u=1$ and no contraction occurs, and $\varepsilon$ is the contraction ratio defined by : $\varepsilon = (l_0 - l/l_0) - l$ is the current muscle length and $l_0$ the initial muscle length. Please note that the contraction ratio will have this same meaning throughout this report for all considered artificial muscles.

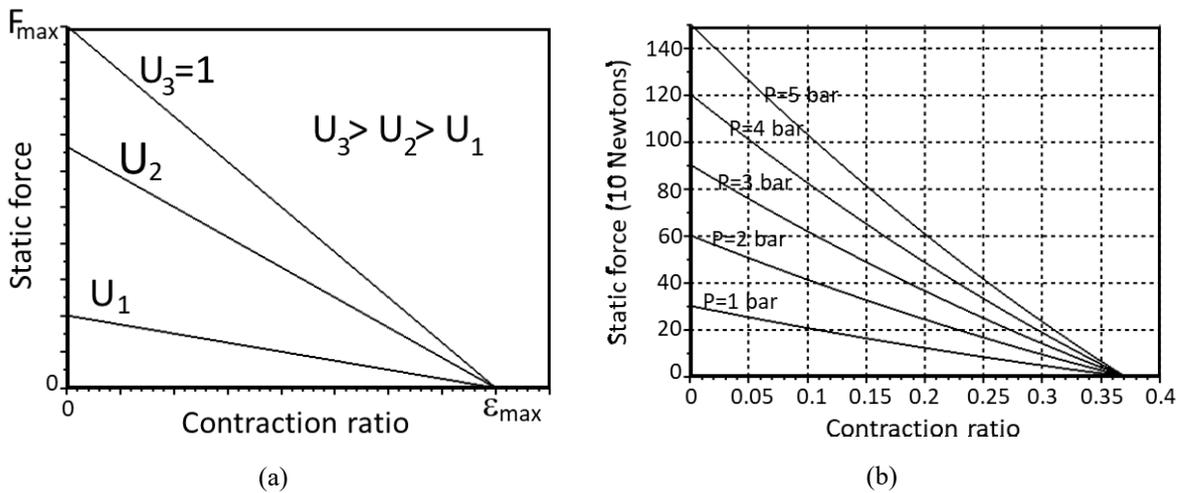

(a)            (b)

**Figure 1.** Interpretation of the theoretical McKibben artificial muscle as a contractile device mimicking the fundamental skeletal muscle model proposed by N. Hogan: (a) Simplified Hogan's model highlighting the variable-stiffness behavior of the skeletal muscle and its constant value for a given neural activation – redrawn from (Hogan, 1984), page 684, (b) Static force of a purely cylindrical McKibben muscle highlighting a stiffness roughly constant during contraction and proportional to the control pressure – see text.

This linear force model can be applied to the pulley-cable antagonistic actuator considered in Fig. 2.a by Hogan as a simplified mechanical model of a physiological joint driven by two antagonist muscles. The following expression of the static torque $T$ results:

$$T = R[F_1(\varepsilon_1, u_1) - F_2(\varepsilon_2, u_2)] \tag{2}$$

where $R$ is the radius of the pulley, $F_1(\varepsilon_1, u_1)$ and $F_2(\varepsilon_2, u_2)$ are the agonist and antagonist forces, according to the superiority of the one over the other. Let us interpret now the scheme of Fig. 2.a proposed by Hogan in the following way: in the initial state – i.e. when joint angle $\theta$ is equal to zero – the two muscles are supposed to be contracted with the same initial contraction ratio $\varepsilon_0$ – typically $\varepsilon_{max}/2$ – while the two muscles are controlled by a same $u_0$-value – typically 0.5 in the case of the normalized $u$-control. When the control values in muscles $u_1$ and $u_2$ are different, and assuming a positive direction when $\varepsilon_1 > \varepsilon_0$, the contraction ratios change according to the following formula:

$$\begin{cases} \varepsilon_1 = \varepsilon_0 + R\theta/l_0 \\ \varepsilon_2 = \varepsilon_0 - R\theta/l_0 \end{cases} \quad (3)$$

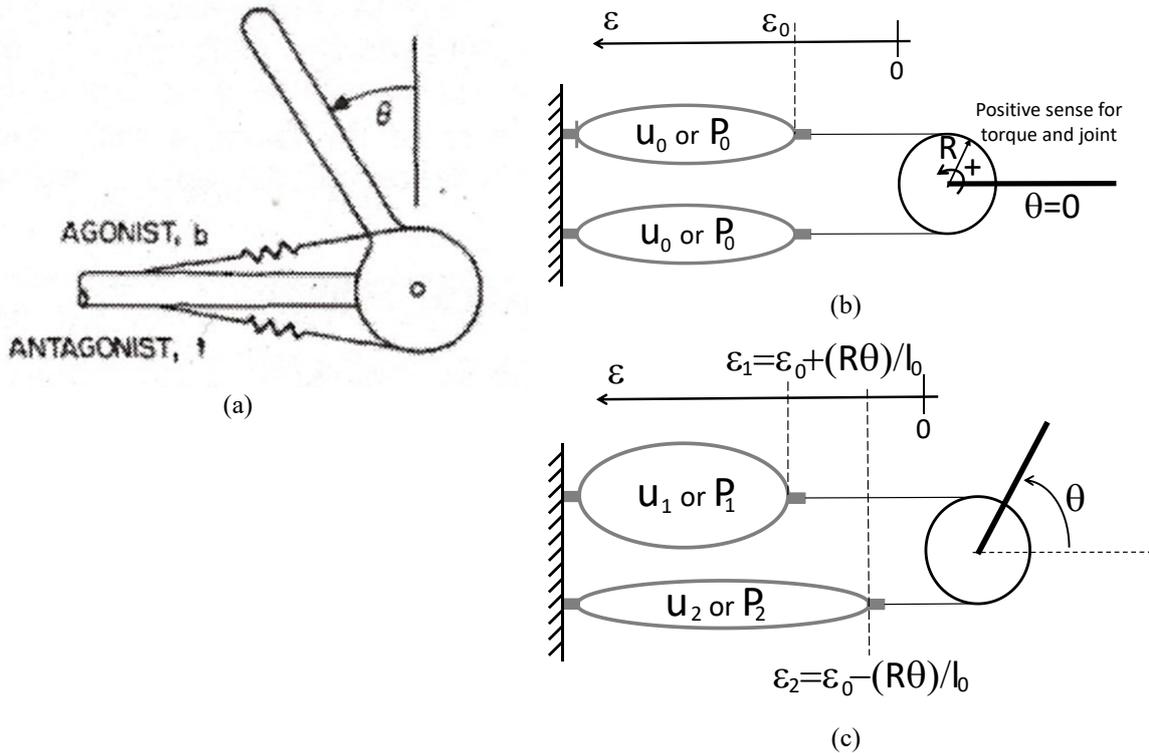

**Figure 2.** Actuator made of two antagonist muscles: (a) Structure proposed by N. Hogan for modelling the typical muscular antagonism in animal body – (Hogan, 1984), page 684, (b) and (c) Initial and current state of the actuator – see text.

Such an actuator model leads to a symmetrical joint range equal to: $[-\left(\frac{l_0}{R}\right)(\varepsilon_{max} - \varepsilon_0), +\left(\frac{l_0}{R}\right)(\varepsilon_{max} - \varepsilon_0)]$. We will limit ourselves to such symmetrical joint range movements in this report – for non-symmetrical joint ranges as imposed for example in the human elbow movement see our study (Tondu et al., 2005). By combining Eqs. (1), (2), (3), we deduce:

$$T = RF_{max}[(1 - \varepsilon_0/\varepsilon_{max})(u_1 - u_2) - (R\theta/l_0\varepsilon_{max})(u_1 + u_2)] \quad (4)$$

and the associated expression of the actuator stiffness $K^1$:

---

[1] Let us insist on the following fact: the skeletal muscle, as any artificial muscle, can be said stable in open-loop (Tondu, 2015), (Tondu, 2019) with respect to some input responsible for a position output which, in our case, is

$$K = -\frac{\delta T}{\delta \theta} = \left(\frac{R^2 F_{max}}{l_0 \varepsilon_{max}}\right)(u_1 + u_2) \tag{5}$$

As a consequence, the actuator stiffness appears to be proportional to the sum $(u_1 + u_2)$ while, for a given stiffness, the relation torque-position is controlled by $(u_1 - u_2)$. In particular, the equilibrium position $\theta_{equ}$ is given by the following relation:

$$\theta_{equ} = \frac{\left(1 - \frac{\varepsilon_0}{\varepsilon_{max}}\right) R F_{max}(u_1 - u_2)}{K} \tag{6}$$

The static model of the purely cylindrical McKibben artificial muscle can be written as follows (Tondu and Lopez, 1995), (Tondu, 2012):

$$F(\varepsilon, P) = (\pi r_0^2) P [a(1-\varepsilon)^2 - b], 0 \leq \varepsilon \leq \varepsilon_{max} = 1 - \sqrt{b/a} \tag{7}$$

with: $a = 3/tan^2(\alpha_0)$ and $b = 1/sin^2(\alpha_0)$ and where $(r_0, l_0, \alpha_0)$ are respectively the initial artificial muscle radius, the initial length radius and the initial braid angle radius. It is important to note that such a static model is purely theoretical in that it combines a set of assumptions difficult to satisfy in practice: a thin-walled inner tube, a cylindrical shape of the artificial muscle throughout the contraction, a constant mechanical solidarity between the inner tube and the textile braid, as no friction during contraction. As it can be seen in Fig. 1.b, the static characteristic of this theoretical McKibben muscle shares several aspects in common with Hogan's model:

- The quasi-linear progression of force with contraction ratio,
- The proportionality of maximum force with pressure, which plays the role of the neural activation,
- The independency of maximum contraction ratio of pressure inducing a stiffness decreasing with pressure like skeletal muscle stiffness decreases with neural activation.

The following equations result for the corresponding MIMO actuator as defined in Figs. 2.b and 2.c where the pressures $P_1$ and $P_2$ are now the control variables in the two antagonist artificial muscles:

$$\begin{cases} T = (\pi r_0^2) R \left[ \left[ a \left((1-\varepsilon_0)^2 + \left(\frac{R\theta}{l_0}\right)^2\right) - b \right](P_1 - P_2) - 2a(1-\varepsilon_0)(R\theta/l_0)(P_1 + P_2) \right] \\ K = (2a(\pi r_0^2) R^2/l_0)[(1-\varepsilon_0)(P_1 + P_2) - \left(\frac{R\theta}{l_0}\right)(P_1 - P_2) \end{cases} \tag{8}$$

It is therefore simple to invert this model for getting, on the one hand, $(P_1 - P_2)$ :

$$\begin{cases} (P_1 - P_2) = (T + K\theta)/(\pi r_0^2) R f(\theta) \quad \text{with:} \\ f(\theta) = a\left((1-\varepsilon_0)^2 - \left(\frac{R\theta}{l_0}\right)^2\right) - b \end{cases} \tag{9}$$

and, on the other hand, $(P_1 + P_2)$:

---

the muscle length or its contraction ratio. In the case of the actuator made of two antagonist muscles, the output is the joint angle $\theta$. When the actuator is slightly deviated from its stable equilibrium position, a return torque $\delta T$ is produced linked to the angle variation $\delta \theta$ by a relation similar to that of a spiral spring: $\delta T = -K\delta\theta$, where $K$ is the actuator stiffness.

$$\begin{cases}(P_1 + P_2) = \left(\frac{1}{A}\right)\left[K + \left(\frac{2aR^2\theta}{l_0^2 f(\theta)}\right)(T + K\theta)\right] \quad \text{with:} \\ A = [2a(\pi r_0^2)R^2(1-\varepsilon_0)]/l_0 \end{cases} \quad (10)$$

For $T = 0$, we can derive the inverse actuator static model giving the couple $(P_1 - P_2, P_1 + P_2)$ with respect to the couple $(\theta_{equ}, K)$ as illustrated in Fig. 3. In order to make a comparison between the McKibben muscle model and the further considered Festo air muscle, we have chosen the following parameters $r_0$=1cm and $\alpha_0$=23.5° leading to a maximum force at $P$=5 bar equal to about 1500 N, which will be the experimental maximum force produced by the considered Festo air muscle at 5 bar. Moreover, we imposed: $\varepsilon_0 = \varepsilon_{max}/2$ i.e. about 0.185. As a consequence, for a chosen initial muscle length chosen equal to 40cm – which will be also the considered length for Festo air muscle – and a pulley-radius equal to 2cm, a maximum joint angle greater than 180° results from the formula: $+\left(\frac{l_0}{R}\right)(\varepsilon_{max} - \varepsilon_0)$. We voluntarily limit the joint angle to: $[-145°, +145°]$. Our simulation of the inverse static actuator model was organized as follows: for a given stiffness, we determine the control pressures $P_1$ and $P_2$ inside the [0, 5bar] range making possible to move totally or partially in the joint range. For the selected geometrical parameters, we illustrated the static actuator functioning in the range [1N.m/rd, 6N.m/rd] with steps of 1 N.m/rd (Fig. 3.a). As it can be seen in Fig. 3.b, stiffness appears to be roughly constant for a given value of $(P_1 + P_2)$ and a bit less roughly proportional to $(P_1 + P_2)$. For a given stiffness, the equilibrium joint angle changes almost linearly with $(P_1 - P_2)$ in accordance with Hogan's model. What happens now if we try to realize such actuator model by means of Festo air muscles?

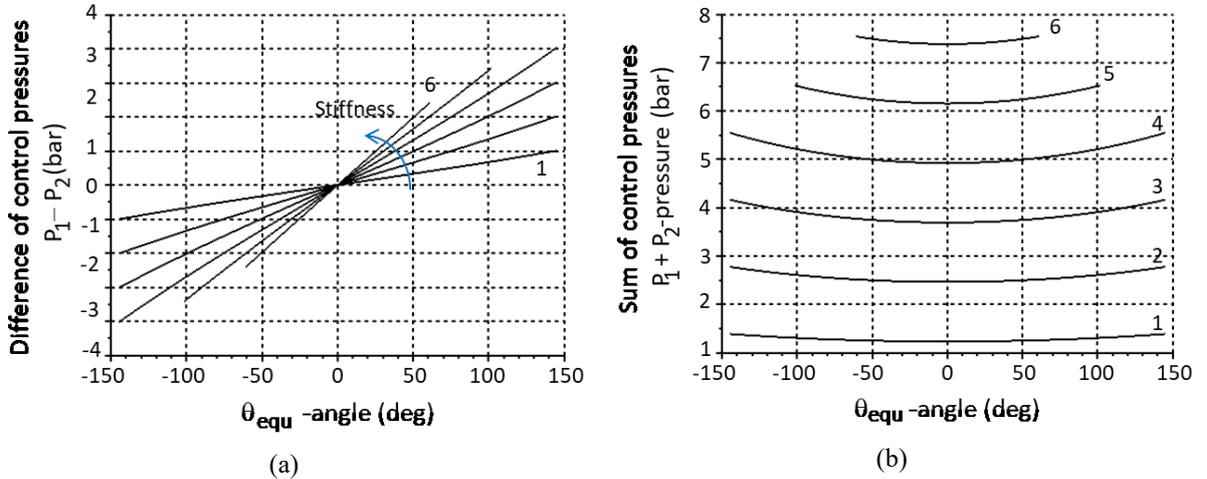

(a)         (b)

**Figure 3.** Simulation of the inverse static model of the theoretical McKibben muscle actuator: Difference (a) and Sum (b) of control pressures for a joint value in the range $[-145°, +145°]$ at constant stiffness varying in the range [1N.m/rd, 6N.m/rd].

## 2. How Festo muscle behaves like a McKibben muscle ?

The static model proposed in Equ. (7) is not realized by the actual versions of McKibben muscle for two practical reasons:

- The non-cylindrical shape during contraction which is a consequence of the non-deformable tips,

- Friction inside the textile braid which prevents the convergence of all constant pressure characteristics towards a same maximum contraction point.

By comparison with hand-made McKibben artificial muscles, the Festo muscle presents the originality of a braided sheath fully integrated into the inner tube[2]. The fact that the braided sheath is outside the inner tube or inside has no consequence because any practical realization of a McKibben muscle, for being efficient, imposes a constant mechanical solidarity between the pressurized tube and the textile braid. Because the contraction of the McKibben contraction is the consequence of the opening of its textile braid, friction between wires, inside the braided sheath, remains. However, we can consider that, whatever the technology used for performing the constant mechanical solidarity between rubber and textile, a McKibben muscle must exhibit what we might think to be its fundamental static property, namely the maximum force proportional to control pressure. Such a physical property directly linked to the theory of McKibben muscle is justified by the fact that, at zero contraction, the assumption of a cylindrical shape is valid and no friction still occurs for reducing this initial force. Curiously, some technological data given by the manufacturer itself do not seem to be in accordance with this fundamental property, as reported in Fig. 5.a. If indeed we consider that the braided sheath is located in the middle of the inner rubber thickness, as illustrated in Fig. 4, the $r_0$-parameter, corresponding to the initial radius of the artificial muscle can be deduced from the knowledge of the internal radius $r_{int}$, given by Festo data sheet, and from the inner tube thickness $t_0$ derived from the direct measurement of the external muscle radius; so we have: $r_0 = r_{int} + t_0/2$. In the case of the MAS-20 muscles, $r_{int}$=1cm and, according to Martens and Boblan (2017), $t_0$ would be equal to about 1.8mm (page 3). As a result, $r_0$ is estimated to about 1.09cm[3]. Moreover, the initial braid angle $\alpha_0$, which is not given by Festo, can be estimated from the knowledge of maximum force, for example at $P$=5 bar or $P$=4bar: in the case of the MAS-20, as reported by the data sheet, we deduce an approximate value of 28.5° for $\alpha_0$.

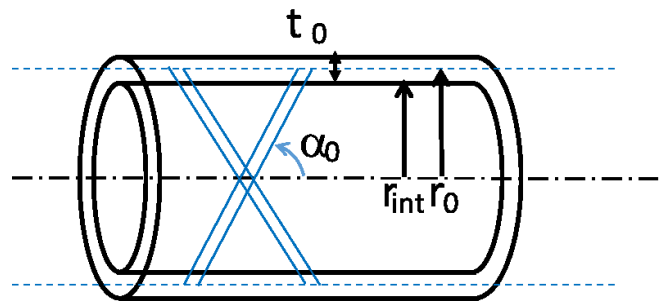

**Figure 4.** Definition of the $r_0$-parameter in the case of the Festo air muscle.

---

[2] At our knowledge, no reference is made in the Festo technological literature to the history of pneumatic artificial muscle as to the product developed by McKibben in the sixties. It is also interesting to note that, in the eighties, the tire manufacturer Bridgestone proposed its own commercial version of the McKibben muscle called *rubbertuator* (E.P.W., 1984), (Inoue, 1988) characterized by an outside braided sheath, before to forgo this product for unknown reasons.

[3] In the framework of this study we did not try to check these data by our own experiments. It is clear however that an accurate analysis of the internal structure of Festo artificial muscles could be very useful particularly with regard to the theory of thin-walled cylinders which is one of the key foundations of the theory of McKibben artificial muscle, as of others inflatable pneumatic artificial muscles. In the case of the Festo muscle considered in this study, the ratio $t_0/r_0$ assumed to be equal to about 0.18 would not respect the thin-walled vessel assumption generally imposing a ratio $t_0/r_0$ less than 0.1.

Lastly, in order to take into consideration the dependence of maximum contraction ratio of the real McKibben artificial muscle with pressure, we have proposed to modify the initial McKibben static force model into the following one (Tondu and Lopez, 2000):

$$F(\varepsilon, P) = (\pi r_0^2)P[a(1 - k(P)\varepsilon)^2 - b], 0 \leq \varepsilon \leq \varepsilon_{max}(P) \qquad (11)$$

where $k(P)$ is an empirical parameter deduced from the reading of the real maximum contraction ratio when $P$ is constant. As shown in Fig. 5.a, the application of this model to the data sheet provided by Festo for its MAS-20 appears to be relatively satisfactorily for $P$=4 and $P$=5 bar, but far less for lower pressures due to an overestimating maximum force at zero contraction. Such a phenomenon, including the absence of contraction for a given 1bar-pressure, could be the consequence of the choice of a too hard rubber or a too thick inner tube. The functioning of a McKibben muscle indeed requires that control pressure be greater than what can be called in rubber theory the "ballooning pressure" (Tondu and Lopez, 1995) which, in the framework of thin-walled rubber tube theory, depends both on the rubber hardness and on the inner tube thickness.

But, in other studies, apparently the same product proposed by the manufacturer shows a clear satisfaction of this McKibben muscle fundamental property, as it can be read on data reported by Sarosi et al. (2015) for a DSMP-20-400N where DSMP seems to be a commercial code for the MAS-20 and 400N means an active length of 400mm – the data relating to the same product reported by Hildebrandt et al. (2005), whose paper includes two Festo's authors, although presented in some manner which makes them not easier to read, seem to be in accordance with those reported by Sarosi et al.. As for the Festo data sheet, we attempted to apply our modified McKibben muscle static model to the available DSMP-20-400N and, as shown on Fig. 5.b, our model is now much more satisfactorily with a clear proportionality between pressure and maximum force at zero-contraction state. Although an $\alpha_0$=25° was better to fit the real data at zero-contraction ratio with maximum force predicted by the model, we applied a little reduced $\alpha_0$ equal to 25.5° in order to be closer to the full contraction curve and so to reduce the effect of initial loss in maximum contraction force resulting, in practice, from the transition between the initial cylindrical shape of the muscle and its spindle shape at tips when contraction starts[4]. Without trying to understand such discrepancy between Festo data, we propose in the framework of this study to develop our analyses from the data reported by Sarosi et al. (2015) justifying to a certain extent the McKibben nature of Festo fluidic muscles. – please note that those data were already reported in (Sarosi and Fabulya, 2012), article that we will discuss later, but in a less clear manner for the reuse of data.

---

[4] An alternative approach consists to add an additional q-parameter, between 0 and 1, modifying Equ. (11) as follows (Andrikopoulos, Nikdakopoulos and Manesis, 2017):
$$F(\varepsilon, P) = (\pi r_0^2)qP[a(1 - k(P)\varepsilon)^2 - b], 0 \leq \varepsilon \leq \varepsilon_{max}(P) \qquad (12)$$
According to the authors: "The role of the introduced q parameter is to decrease the resulting curve enough to compensate for the resisting forces generated via friction phenomena, the non-zero thickness of the elastic tube and the bearing effect, which are not taken into account in the fundamental model" (page 2649).

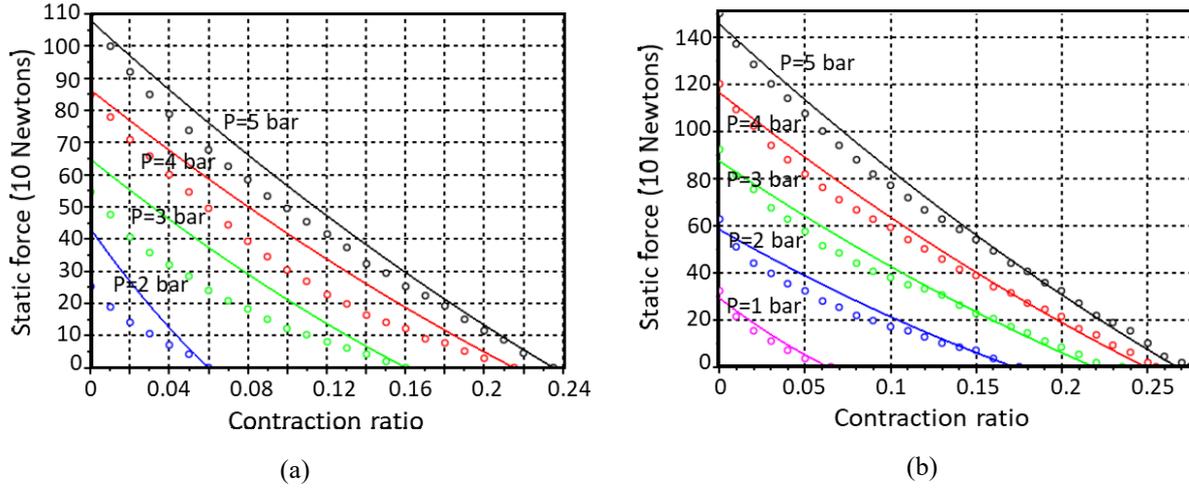

**Figure 5.** Attempt to apply a McKibben static force model (in full line) to Festo air muscle: (a) Case of the Festo data sheet MAS-20 highlighting the lack of proportionality between maximum force and pressure – data given by Festo (2004), (b) Case of a DSMP-20-400N – data reported by Sarosi et al. (2015) – see text.

By comparison with other hand-made McKibben artificial muscles (Caldwell, Medrani-Cerda and Goodwin, 1995), (Tondu and Lopez, 2000), one tie however appears to be common to all available data relating to static force developed by Festo muscles: the relatively weak maximum contraction ratios when pressure is low, especially around 1 bar. As a result, it is not so easy to propose a phenomenological model of static force in the typically working range [0-5bar] considered in this study. Such models have already been proposed, as particularly well synthetized by Martens and Boblan (2017), including a recent model proposed by these authors. The actual diversity of these models highlights the difficulty to exhibit a model combining a physical background with good accuracy. Moreover, because our aim consists in deriving, from a chosen muscle static force model, a static actuator model, and then inverting it in a simple way, the muscle model much be as simple as possible. Starting from theoretical McKibben muscle model, we could consider the new following model:

$$F(\varepsilon, P) = (\pi r_0^2)[(a-b)P - \varepsilon f(\varepsilon, P)] \quad (13)$$

where the first term expresses the initial maximum force deduced from the knowledge of $r_0$ and $\alpha_0$-parametrs and the second term attempts to capture the shape of the curve between the initial zero contraction point and maximum contraction point at zero force. Due to the presence of the pressure $P$ in the first term, responsible for the fundamental property of maximum force proportional with pressure, it may seem like a good idea to limit the function $f(\varepsilon, P)$ to only one $\varepsilon$-dependent function; in the case of a polynomial function, we can write:

$$f(\varepsilon, P) = f(\varepsilon) = a_0 + a_1\varepsilon + a_2\varepsilon^2 + \cdots \quad (14)$$

A simple way for determining the $a_i$-coefficients can consist to derive them from the knowledge of maximum contraction ratio for a set of constant $P$-characteristics: if $n$ $\varepsilon_{max}(P_i), 1 \leq i \leq n$ are given, a $n$-order polynomial $f(\varepsilon)$ results whose $(a_i)_{1 \leq i \leq n}$-coefficients are the solutions of the following matrix equation:

$$\begin{bmatrix} \varepsilon_{max}(P_1) & \varepsilon_{max}^2(P_1) & \varepsilon_{max}^3(P_1) & \cdots \\ \varepsilon_{max}(P_2) & \varepsilon_{max}^2(P_2) & \varepsilon_{max}^3(P_2) & \cdots \\ \vdots & \vdots & \vdots & \end{bmatrix} \begin{bmatrix} a_0 \\ a_1 \\ \vdots \end{bmatrix} = (a-b) \begin{bmatrix} P_1 \\ P_2 \\ \vdots \end{bmatrix} \quad (15)$$

We applied this approach to the data of our DSMP-20-400N Festo muscle: while a 5th-order polynomial leads to a deceptive result, the best result is got for a 3rd-order polynomial with $(P_1, P_2, P_3) = (3,4,5)$ bar, as illustrated in Fig. 6. The well-known "wandering" phenomenon peculiar to spline interpolation is the obvious cause, and it is clear that more sophisticated interpolation methods considering the full data of the experimental curves are necessary.

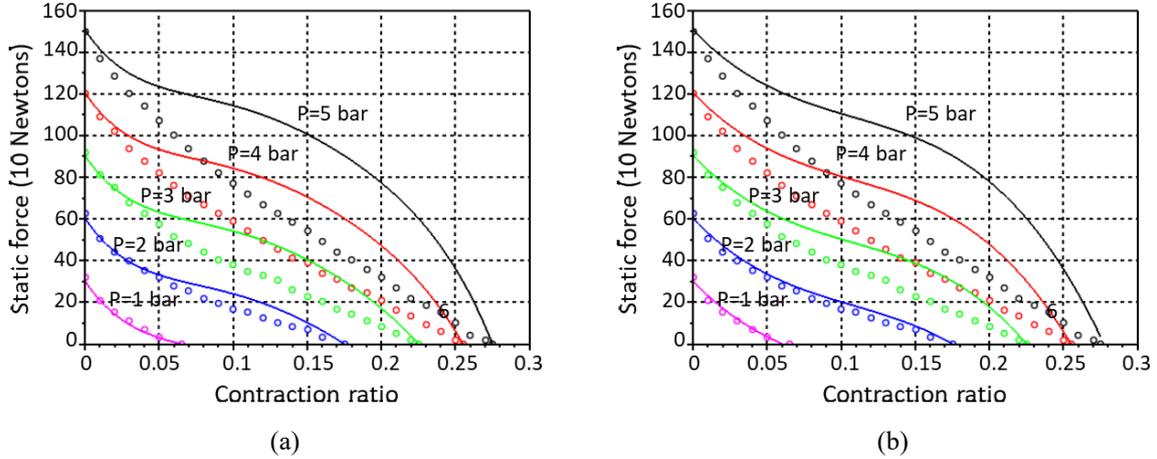

(a)          (b)

**Figure 6.** Attempt to use a polynomial function for interpolating experimental data between first zero-contraction point and final zero-force point: (a) Fifth-order polynomial, (a) Third-order polynomial.

Dropping all reference to the McKibben artificial muscle, Hildebrandt et al. (2005) – including two members of Festo company – proposed an original model in the form, rewritten with our own notations:

$$F_{Hildebrandt}(\varepsilon, P) = f_1(\varepsilon)P - f_2(\varepsilon) = (c_0 + c_1\varepsilon + c_2\varepsilon^2)P - (d_0 + d_1\varepsilon + d_2\varepsilon^2 + d_3\varepsilon^3 + d_4\varepsilon^{2/3}) \qquad (16)$$

Let us first remark that such a model is in accordance with the general definition of an artificial muscle (Tondu, 2015), (Tondu, 2019) whose equilibrium position results from two antagonist effects – here the term $f_1(\varepsilon)P$ that the authors interpret as a virtual piston and the term $f_2(\varepsilon)$ playing the role of a counter force. By comparison with alternative models, like that proposed by Wickramatunge et al. (2010) involving a squared pressure or that proposed by Takosoglu et al. (2016) involving an exponential function with pressure, Hildebrandt's and coauthors' model offers the great advantage, due to its single term in $P$, of being easily applicable for the determination of an inverse static model of the antagonist muscle actuator[5]. However, the presence of the final term '$d_4\varepsilon^{2/3}$' emphasizes the difficulty to get, even with a respectable number of parameters, an accurate model on the basis of polynomial functions in $\varepsilon$: as reported by the authors, "To map the strong rise of the force at small contraction displacements precisely, it would be necessary to use a polynomial function of the 30th order for the spring force-behavior. To reduce the order, a polynomial of the 3rd order is supposed which is added by a power function" (page 682). Hildebrandt's and coauthors' model also causes a questionable effect: the stiffness of the artificial muscle approaches infinity at zero contraction. Even if it is clear that the stiffness of any McKibben muscle-type including Festo fluidic muscle is at its higher value when contraction ratio is equal to zero, one can ask whether an alternative model with no infinite stiffness is possible. Another very interesting purely mathematical model is that proposed by Sarosi and Fabulya (2012), which can be written as follows:

$$F_{Sarosi}(\varepsilon, P) = f_1(\varepsilon)P - f_2(\varepsilon) = (c_1 exp^{c_6\varepsilon} + c_2\varepsilon + c_3)P - (c_4 exp^{c_6\varepsilon} - c_5) \quad (17)$$

---

[5] In their new approach, Martens and Boblan (2017) also propose an interesting static model where only a factor in pressure is considered based on a physical study of elastic forces acting inside the inner rubber tube. However, this model, as far as we understood it, is based on the assumption of a purely cylindrical shape for the inner tube. According to the authors, its accuracy would be greater than all other available models. The fact also that only simulation data are reported in the article and no curve is shown makes it difficult to well appreciate its quality.

This 6-parameter model has a similar structure than the previous one but with fewer parameters. As in Hildebrandt's and coauthors' model, the artificial muscle stiffness can easily be deduced in a closed form (Sarosi et al., 2015). However the presence of the exponential term '$exp^{c_6\varepsilon}$' leads to complex formula in the case of the antagonist muscle actuator, not considered by the authors.

Without downplaying the interest of those powerful models, we tried to propose a simpler and so efficient model. Coming back to our Equ. (13), we propose to consider the following $f(\varepsilon, P)$–function:

$$f(\varepsilon, P) = \frac{cP + e}{P + d} \quad (18)$$

where ($c$, $d$, $e$) are three real parameters. It is easy to show that such a function requires for determining the inverse antagonist muscle actuator to solve a third-order polynomial equation in $P_1$, which can be made in a closed form, as we will illustrate it further. The ($d$, $e$)-parameters can be deduced from the data of two maximum contraction ratios $\varepsilon_{max}(P_I)$ and $\varepsilon_{max}(P_{II})$ associated respectively to the constant pressure $P_I$ and $P_{II}$ characteristics as follows:

$$\begin{cases} d = -\frac{P_{II}^2 - \left(\frac{\varepsilon_{max}(P_{II})}{\varepsilon_{max}(P_I)}\right)P_I^2}{\left(P_{II} - \left(\frac{\varepsilon_{max}(P_{II})}{\varepsilon_{max}(P_I)}\right)P_I\right)} + \frac{c}{(a-b)} \frac{(P_{II} - P_I)\varepsilon_{max}(P_{II})}{\left(P_{II} - \left(\frac{\varepsilon_{max}(P_{II})}{\varepsilon_{max}(P_I)}\right)P_I\right)} \\ e = \frac{(a-b)P_I(P_I + d)}{\varepsilon_{max}(P_I)} - cP_I \end{cases} \quad (19)$$

The resting $c$-parameter can then be used for getting the best interpolation for other considered $P_i$-characteristics with $i \neq I$ and $i \neq II$. In the case of our considered Festo muscle, the best result was got for $(P_I, P_{II}) = (3bar, 5bar)$ and $(\varepsilon_{max}(P_I), \varepsilon_{max}(P_{II})) = (0.225, 0.275)$: $d = -10.5$bar, $e = -779$bar$^2$, with $c = 0$. The resulting static model is shown in dashed line in Fig. 7.a near the original model of Equ. (11) in full line: it appears to be relatively satisfactorily as a simplified model for a further feed-forward control; it however overestimates the force at $P$=1bar when contraction ratio increases. It is interesting to note that a much better adherence would be obtained by considering a double set of ($d,e$)-parameters for lower pressures, on the one hand, and for upper pressures, on the other hand, as done by Wickramatunge and Leephakpreeda (2010) with their stiffness-based model:

$$\begin{cases} F_{Wickramatunge} = K_M(l_s, P)l_s \\ K_M(l_s, P) = c_3P^2 + c_2Pl_s + c_1l_s^2 + c_0 \end{cases} \quad (20)$$

where $l_s$ is the "stretched length" of the muscle i.e. the variation of the muscle length with respect to its minimum length: $l_s = l - \min(l)$. To be relevant, the model proposed by the authors considers – for elongation and contraction to take into account the hysteresis phenomenon – two sets of $c_i$- parameters ($i$=0 to 3) for the range [0-2.5bar] and for the range [2.5-5bar] whose values can clearly differ, especially for the term in $P^2$. Such a partitioning of identified parameters is adequate for closed-loop control; this is the reason why we propose to reject it. Moreover, by comparison with Wickramatunge and Leephakpreeda' model, our proposed simplified static force model induces the following expression of the artificial muscle stiffness $K_M$ as follows:

$$K_M(l,P) = -\frac{\delta F(l,P)}{\delta l} = +\frac{\delta F(\varepsilon,P)}{l_0 \delta \varepsilon} = [(\pi r_0^2)/l_0]\frac{cP+e}{P+d} \quad (21)$$

which is so constant for a given $P$-control pressure, as can be checked on Fig. 7.b. It is clear that the proposed simplified model does not correctly evaluate the stiffness near the zero-contraction state i.e. when the static contraction force drastically changes. In the case of our antagonist muscle actuator, we however can assume to limit the actuator joint range to angle values until a minimum contraction ratio corresponding to a given threshold, $\varepsilon_{threshold}$, for example 0.025%. Under this assumption and whatever the chosen artificial muscle model, muscle stiffness could be considered as constant with pressure.

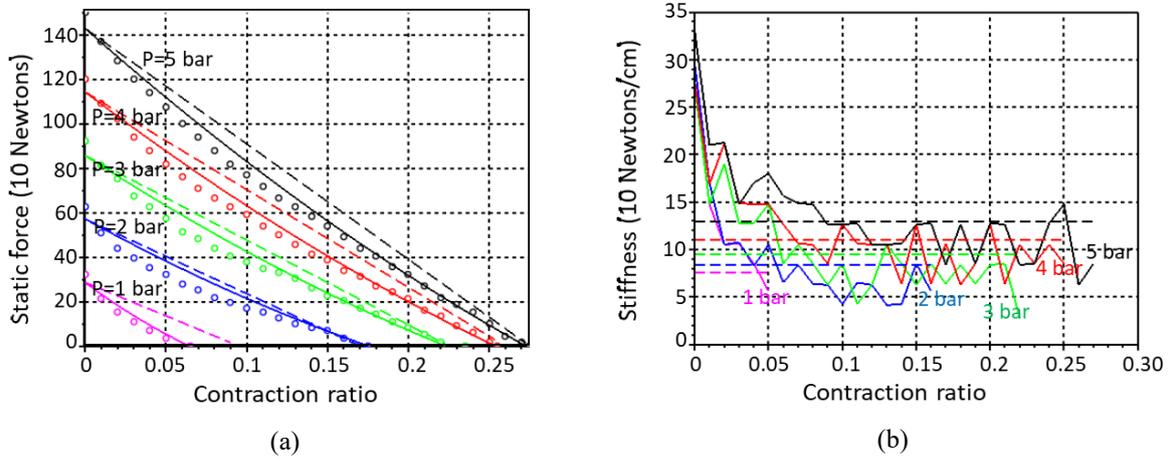

(a)        (b)

**Figure 7.** Comparison between the real data for the DSMP-20-400N Festo air muscle, the proposed McKibben-type model given in Equ. (11) – in solid line, and the new proposed model given in Equs. (13) and (18) – in dashed line (a), and Comparison between the model stiffness – in dashed line – and the estimated Festo muscle stiffness – in full line (b).

## 3. Direct and inverse static model of the antagonist actuator

On the same terms as before, in the case of our theoretical McKibben muscle actuator, we derive the following direct torque-stiffness model for the Festo actuator:

$$\begin{cases} T = (\pi r_0^2)R[(a-b)(P_1-P_2) - e\varepsilon_0\left(\frac{1}{P_1+d} - \frac{1}{P_2+d}\right) - \left(\frac{eR\theta}{l_0}\right)\left(\frac{1}{P_1+d} + \frac{1}{P_2+d}\right)] \\ K = A\left(\frac{1}{P_1+d} + \frac{1}{P_2+d}\right) \quad \text{with } A = \frac{(\pi r_0^2)R^2 e}{l_0} \end{cases} \quad (22)$$

Let us also note that we can exhibit the following direct model for the equilibrium joint angle, while it is much more difficult for more complex models as those proposed by Hildebrands et al. (2005) or Sarosi and Fabulya (2012):

$$\theta_{equ} = \left(\frac{l_0}{R}\right)\frac{(P_1-P_2)}{(P_1+P_2+2d)}[\varepsilon_0 + (a-b)\frac{(P_1+d)(P_2+d)}{e}] \quad (23)$$

We assume that the control pressures must keep in the range [0, 5bar], and we propose to select the initial contraction ratio according to the following formula:

$$\varepsilon_0 = \frac{\varepsilon_{max}(5bar)}{2} \qquad (24)$$

For the considered Festo muscle, we get, from the experimental value $\varepsilon_{max}(5bar) = 0.275$, an $\varepsilon_0$-value equal to 0.1375. Moreover, the maximum joint angle can be determined according to the following formula:

$$\theta_{max} = -\theta_{min} = (l_0/R)(\varepsilon_0 - \varepsilon_{threshold}) \qquad (25)$$

which gives, for $l_0 = 40cm$ and $R = 2cm$, with a choice for $\varepsilon_{threshold}$ equal to 0.025: $\theta_{max} \sim 129°$. We will chose $\theta \in [-125°, +125°]$ in simulations shown in Fig. 8.

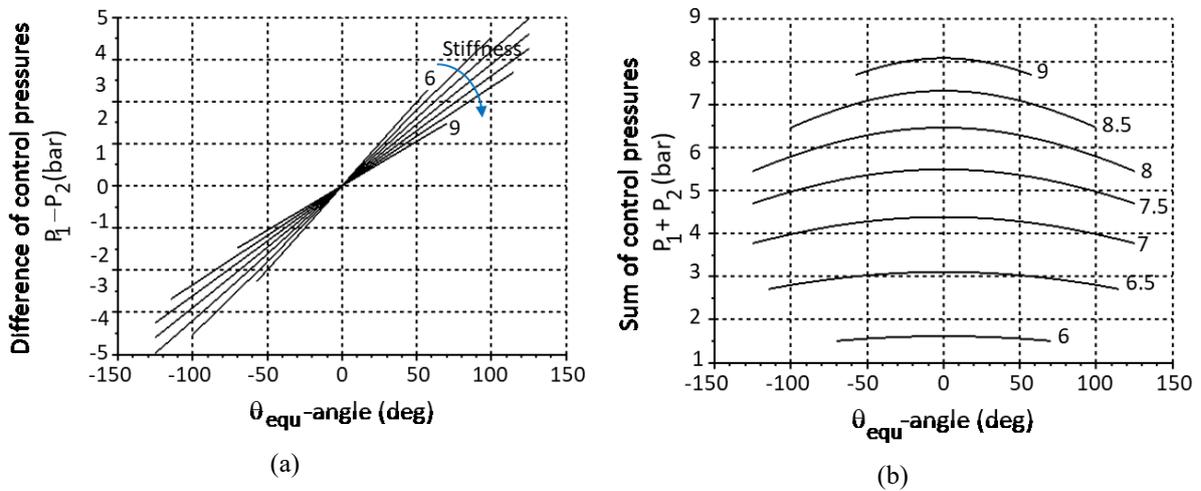

(a)  (b)

**Figure 8.** Simulation of the inverse static model of the Festo air muscle actuator: Difference (a) and Sum (b) of control pressures in the antagonist artificial muscles for a joint value in the range [−125°,+125°] at constant stiffness varying in the range [6 N.m/rd ,9 N.m/rd].

By comparison with the theoretical McKibben muscle actuator, the following points can be highlighted:

- Due to limited contraction ratio at low pressure, for a same pulley radius, joint range is smaller and is even smaller by comparison with other hand-made McKibben artificial muscle actuators – see, for example, (Tondu and Lopez, 2000),
- Static position characteristic at constant stiffness still appears to be roughly linear with respect to $(P_1 - P_2)$,
- At same maximum force performance, Festo muscle actuator appears to be stiffer. This is a consequence of limited maximum contraction ratios but, despite the fact of relatively large variations in those maximum contraction ratios, the Festo muscle actuator, as we attempted to model it, exhibits a stiffness varying relatively little with respect to $(P_1 + P_2)$. It is interesting to note that, while the theoretical McKibben muscle actuators exhibits slightly concave $(P_1 + P_2)$-characteristics versus joint position, at constant stiffness, the Festo air muscle actuator exhibits larger convex similar characteristics due to the fact that, while stiffness clearly decreases with pressure in the case of the McKibben muscle, this property is well less pronounced in the case of the Festo muscle.

# Conclusion

The Festo air muscle is undoubtedly the most used commercial version of the McKibben artificial muscle. By comparison with other hand-made McKibben muscles, the Festo air muscle is considered a robust device. In the framework of this study, we did not try to characterize the physical properties, which could explain such a peculiar robustness by comparison with other hand-made McKibben muscles. We however can suggest that the limited maximum contraction ratios at low pressure could be the consequence of a relatively thicker and/or stronger rubber, responsible of the Festo air muscle robust character. Such relatively large variations in maximum contraction ratios with control pressure in a typical [0-5bar] range has a surprising consequence: the Festo muscle stiffness, when it went over the zero-contraction area, varies relatively little by pressure. This property seems to turn away the Festo air muscle from theoretical McKibben artificial muscle whose stiffness clearly diminishes when pressure decreases in accordance with the fundamental Hogan's model.

However, as we attempted to show it, an actuator made of two antagonist Festo air muscles is able to be a relevant hybrid position-stiffness MIMO system controlled by the two pressures inside the artificial muscles. Because stiffness cannot be directly measured by a sensor, an accurate model of the actuator is necessary for estimating it while joint position is controlled in open-loop or in closed-loop. Such an actuator model must be derived from a static model of the artificial muscle, the most accurate possible but also moderately complex in order to make possible a closed form inverse actuator model mandatory for stiffness evaluation. If the Festo air muscle clearly appears as a practical McKibben artificial muscle in the sense it can be in accordance with the fundamental property of a zero-contraction maximum force proportional to pressure, its large variation of maximum contraction ratio with pressure seems to be a factor making difficult to exhibit a simple static force model relevant in the practical [0-5bar] pressure range.

The proposed simplified static force model for the Festo air muscle based on the McKibben artificial muscle theory includes an empirical term in $'\varepsilon \frac{cP+e}{P+d}'$ where ($c$, $d$, $e$) are three parameters supposed to be able to catch the diversity of maximum contraction ratio with control pressure $P$. A closed form solution results for the inverse actuator model which only requires to solve a third-order polynomial equation. As shown by our simulations, if it cannot be said that the actuator stiffness is proportional to the sum of muscle control pressures, this last one varies relatively little for a given stiffness throughout the joint range as long as control pressures keep in the imposed [0-5bar] joint range. The proposed static force model fails however to be accurate at low pressure, typically around 1 bar pressure. Further work will try to better analyze the relevance of purely mathematical models, especially combining polynomial functions in contraction ratio with a *P*-term, as initiated by Festo's engineers themselves.

# References :

Andrikopoulos G., Nikolakopoulos G., Manesis S., "Novel Consideration on Static Force Modeling of Pneumatic Muscle Actuators", *IEEE/ASME Trans. Mechatronics*, 21, 2016, pp. 2647-2659.

Caldwell D.G., Medrano-Cerda G.A. and Goodwin M., "Control of Pneumatic Muscle Actuators", *IEEE Control Systems Magazine*, 15(1), 1995, pp. 40-48.

Chou C.-P. and Hannaford B., "Measurement and Modelling of McKibben Pneumatic Artificial Muscles", *IEEE Trans. on Robotics and Automation*, 12(1), 1996, pp. 90-102.

E.P.W., "Rubber Muscles Take Robotics one Step Further", *Rubber Development*, 37(4), 1984, pp. 117-119.

Festo AG, Datasheet–Fluidic Muscle DSMP/MAS, Festo, Esslingen am Neckar, Germany, November 2004.

Hildebrandt A., Sawodny O., Neumann R. and Hartmann A., "Cascaded Control Concept of a Robot with two Degrees of Freedom Driven by four Artificial Pneumatic Muscle Actuators", *Proc. of the 2005 American Control Conference*, Portland, OR, USA, June 2005, pp. 680-685.

Hogan N., "Adaptive Control of Mechanical Impedance by Coactivation of Antagonist Muscles", *IEEE Trans. on Automatic Control*, vol. AC-29, n°8, August 1984, pp. 681-690.

Inoue K., "Rubbertuators and Applications for Robots", *Proc. of the 4$^{th}$ Int. Symp. on Robotics Research*, Cambridge, MA, USA, 1988, pp. 57-63.

Martens, M. and Boblan I., "Modeling the Static Force of a Festo Pneumatic Muscle Actuator: A New Approach and a Comparison to Existing Models", *Actuators*, 6(33), 2017 (11 pages).

Sarosi J. and Fabulya, "Mathematical Analysis of the Function Approximation for the Force Generated by Pneumatic Artificial Muscle", *Trans. on Mechanics, Scientific Bulletin of the Politechnica, University of Timisoara*, 57(71), 2012, pp. 59-64.

Sarosi J., Biro I., Nemeth J., Cveticanin L., "Dynamic Modeling of a Pneumatic Muscle Actuator with two-Direction Motion", *Mechanism and Machine Theory*, 85, 2015, pp. 25-34.

Takosoglu J.E., Laski P.A., Blasiak S., Bracha G. and Pietrala D., "Determining the Static Characteristics of Pneumatic Muscles", *Measurement and Control*, 49(2), 2016, pp. 62-71.

Tondu B. and P. Lopez, "Theory of an Artificial Pneumatic Muscle and Application to the Modelling of McKibben Artificial Muscle", *C.R.A.S., French National Academy of Sciences*, Series IIb, 320, 1995, pp. 105-114 (In French with an abridged English version).

Tondu B. and P. Lopez, "Modeling and Control of McKibben Artificial Muscle Robot Actuators", *IEEE Control Systems Magazine*, 20(2), 2000, pp. 15-38.

Tondu B., Ippolito S., Guiochet J. and Daidie A., "A Seven-Degrees-of-Freedom Robot-Arm Driven by Pneumatic Artificial Muscles for Humanoid Robots", *The Int. J. of Robotics Research*, 24(4), 2005, pp. 257-274.

Tondu B., "Modelling of the McKibben Artificial Muscle: A Review", *J. of Intelligent Material Systems and Structures*, 23(3), 2012, pp. 225-253.

Tondu B., "What is an Artificial Muscle? A Systemic Approach", *Actuators*, 4(4), 2015, pp. 336-352.

Tondu B., "Towards a Theory of Actuators: A New Classification Proposal for Actuation Systems", *Sensors and Actuators A: Physical*, 289, 2019, pp. 108-117.

Wickramatunge K.C. and Leephakpreeda T., "Study on Mechanical Behaviors of Pneumatic Artificial Muscle", *Int. J. of Engineering Science*, 48, 2010, pp. 188-198.

## Annex: Closed form solution of a third-order polynomial equation

Let us consider the equation:
$$x^3 + a_2 x^2 + a_1 x + a_0 = 0$$

1. The original equation can be put in the following form:
$$y^3 + py + q = 0$$

with: $\begin{cases} y = x + \frac{a_2}{3} \\ p = a_1 - \left(\frac{a_2^2}{3}\right) \\ q = a_0 - \left(\frac{a_1 a_2}{3}\right) + \left(\frac{2 a_2^3}{27}\right) \end{cases}$

2. Let us put: $D = q^2/4 + p^3/27$.

If D>0, one sole real solution exists:

$$x_{sol} = \left(\sqrt[3]{-\left(\frac{q}{2}\right) + \sqrt{D}}\right) + \left(\sqrt[3]{-\left(\frac{q}{2}\right) - \sqrt{D}}\right) - \left(\frac{a_2}{3}\right)$$

If D<0, three real solutions exist:

$$\begin{cases} x_{sol1} = 2\sqrt{-\frac{p}{3}} \cos\left(\frac{t}{3}\right) - \frac{a_2}{3} \\ x_{sol2} = 2\sqrt{-\frac{p}{3}} \cos\left(\frac{t}{3} + \frac{2\pi}{3}\right) - \frac{a_2}{3} \\ x_{sol3} = 2\sqrt{-\frac{p}{3}} \cos\left(\frac{t}{3} + \frac{4\pi}{3}\right) - \frac{a_2}{3} \end{cases}$$

with: $t = arcos(-\frac{q}{2r})$, $t \epsilon [0, 2\pi]$ and $r = \sqrt{-p^3/27}$.

For practical reasons, the case D=0 is not considered here.

For more details, the French reader can consult the free on-line synthetic presentation of Serge Mehl: "ChronoMath, une chronologie des Mathématiques", serge.mehl.free.fr, 1996.